\pdfoutput=1

\documentclass[11pt]{article}

\usepackage{ACL2023}

\usepackage{times}
\usepackage{latexsym}
\usepackage{graphicx}
\usepackage{url}

\usepackage[T1]{fontenc}

\usepackage[utf8]{inputenc}

\usepackage{microtype}

\usepackage{inconsolata}
\usepackage[symbol]{footmisc}

\title{Large Language Models Are Partially Primed in Pronoun Interpretation}

\author{Suet-Ying Lam$^{\dag}$\thanks{$^{\ast}$Equal contribution by alphabetical order. Correspondence to \texttt{qcz@u.northwestern.edu}} \quad Qingcheng Zeng $^{\ddag\ast}$ \quad Kexun Zhang $^{\diamondsuit\ast}$ \quad Chenyu You $^{\clubsuit}$ \quad Rob Voigt$^{\ddag}$ \\ \\
$^{\dag}$ Department of Linguistics, UMass Amherst \\ $^{\ddag}$ Department of Linguistics, Northwestern University \\  $^{\diamondsuit}$ Department of Computer Science, UCSB\\$^{\clubsuit}$ Department of Electrical Engineering, Yale University} 

\begin{document}
\maketitle
\begin{abstract}
While a large body of literature suggests that large language models (LLMs) acquire rich linguistic representations, little is known about whether they adapt to linguistic biases in a human-like way. The present study probes this question by asking whether LLMs display human-like referential biases using stimuli and procedures from real psycholinguistic experiments. Recent psycholinguistic studies suggest that humans adapt their referential biases with recent exposure to referential patterns; closely replicating three relevant psycholinguistic experiments from \citet{johnson2022frequency} in an in-context learning (ICL) framework, we found that InstructGPT adapts its pronominal interpretations in response to the frequency of referential patterns in the local discourse, though in a limited fashion: adaptation was only observed relative to syntactic but not semantic biases. By contrast, FLAN-UL2 fails to generate meaningful patterns. Our results provide further evidence that contemporary LLMs discourse representations are sensitive to syntactic patterns in the local context but less so to semantic patterns. Our data and code are available at \url{https://github.com/zkx06111/llm_priming}.
\end{abstract}

\section{Introduction}
While neural network models, and particularly pre-trained large language models 
have shown excellent performance at particular language processing tasks, many questions remain about the extent to which models optimized for such performance encode, as a side effect, human-like linguistic knowledge and cognitive biases.
We know that they do to some extent; existing work has shown, for example, that neural models encode aspects of human-like long-distance number agreement \cite{gulordava-etal-2018-colorless}, incremental syntactic state \cite{futrell-etal-2019-neural}, and syntactic generalization more broadly \cite{hu-etal-2020-systematic}. 
In this paper, we examine whether FLAN-UL2 \cite{tay2023ul2} and InstructGPT \cite{https://doi.org/10.48550/arxiv.2203.02155}, two representative LLMs, display adaptation in pronoun interpretation when exposed to consistent referential patterns in the local discourse context.

Compared with syntactic or lexical knowledge, representing referential knowledge is possibly more complex; we know from psycholinguistic studies that human referential interpretation integrates multiple levels of linguistic structure.
Humans do not interpret ambiguous pronouns at random but are guided by both syntactic and semantic information. It is well-established that absent other cues humans prefer a syntactic subject in choosing the antecedent of the ambiguous pronoun, i.e., \textbf{subject bias} \cite{ariel1990accessing, brennan1995centering}. In example (1), \textit{she} is more likely to be interpreted as the subject \textit{Ada} than the non-subject \textit{Eva} \footnote[5]{Indices of pronouns in examples indicate the preferred referent.}, even though both referents are possible antecedents for the pronoun. 
\begin{enumerate}
    \item [(1)] Ada$_1$ talked with Eva$_2$. She$_1$...
\end{enumerate}

People are also sensitive to the semantic structure of the sentence when choosing an antecedent for an ambiguous pronoun, in addition to syntactic information. In a transfer event that depicts a transfer-of-possession from one entity (the source) to another (the goal), they prefer the goal referent (Ada in (2), Eva in (3)) over the source referent (Eva in (2), Ada in (3)) to be the antecedent \cite{arnold2001effect,arnold1998reference}, i.e., \textbf{goal bias}:
\begin{enumerate}
    \item[(2)] Goal-source (gs) verb: \\Ada$_1$ received a letter from Eva$_2$. She$_{1}$...  
    \item[(3)] Source-goal (sg) verb: \\Ada$_1$ sent a letter to Eva$_2$. She$_{2}$... 
\end{enumerate}

In sum, people exhibit sensitivity to both syntax (subject bias) and semantics (goal bias) during pronoun interpretation. Importantly, these levels of linguistic structure are frequently entwined since both can influence referential interpretation. At times these influences may push in different directions: for instance, \textit{Ada} in [3] is both the syntactic subject and the semantic source. 


Building on a long tradition investigating preferences in pronoun interpretation, recent psycholinguistic studies have probed into the deeper question of the origin of these biases. 
One hypothesis is that referential biases come from linguistic experience: when a bias occurs very frequently, people will tend to adapt this more frequent referential pattern, both in the immediate exposure as well as more large-scale past experience \cite{arnold1998reference, arnold2001effect}. Recent evidence has provided support to this idea by demonstrating that recent exposure to certain referential patterns did change people's referential biases. In a series of psycholinguistic studies, \citet{johnson2022frequency} show that after reading numerous stories that consistently show a particular referential bias, e.g., always referring to the non-subject or source referent, people did have a stronger preference for these primed referents.


Given this line of psycholinguistic research, the current study investigates the extent to which LLMs adapt and vary referential biases in pronoun interpretation through exposure to referential patterns in the local context.
To do so we replicated actual psycholinguistic experiments from \citet{johnson2022frequency} in LLMs using an ICL framework and asked whether the responses of LLMs display adaptation from exposure to referential patterns like human experimental participants. Comparing syntactically-motivated to semantically-motivated exposure conditions will allow us to first examine whether LLMs display human-like subject bias and goal bias and further understand the extent to which LLMs make use of local frequency changes in discourse representations in these categories.

In-context learning refers to LLMs' ability to learn from demonstrations written in natural language prompts. Compared to previous work that has largely examined the encoding of such discourse knowledge using zero-shot inference \cite{upadhye-etal-2020-predicting}, ICL is particularly suitable for experimental simulation since it replicates the naturalistic context in which later responses draw upon exposure to previous examples.


Foreshadowing our results, we find that InstructGPT can adapt and thus vary its syntactic bias from exposure to referential patterns in the local context, but the same is not true for semantic bias. Given that InstructGPT still exhibits a goal bias in spite of local discourse priming, we argue this suggests LLMs only encode partial semantic knowledge in referential processing. To sum up, our contributions can be summarized as follows:
\begin{itemize}
    \item [1)] We extended a discourse understanding evaluation to state-of-the-art LLMs from a new perspective, asking whether LLM's referential bias can be modified by exposure to particular referential patterns, like how humans adapt referential bias from experience.
    \item [2)] To the best of our knowledge, we are the first study that replicates actual psycholinguistic experiments using the ICL framework and compares LLMs' behaviors with real human participants.
    \item [3)] We present results in this context showing further evidence that InstructGPT can acquire abstract syntactic knowledge in referential interpretation to some extent, but not semantic knowledge.
\end{itemize}

\section{Related Work}
A growing body of literature has suggested that LLMs encode rich representations of linguistic structure at various levels, including aspects of syntax, semantics, and reference encoded throughout their representations  \cite{tenney2019rediscovers}. 
One of the most well-documented lines of this work demonstrates that these models can acquire diverse elements of syntactic knowledge \cite{gulordava-etal-2018-colorless,futrell-etal-2019-neural,hu-etal-2020-systematic}.

This capacity for encoding linguistic understanding extends to priming effects with psycholinguistic analogies to humans. At the syntactic level, \citet{10.1162/tacl_a_00504} explored structural priming in various autoregressive LLMs and found priming effects despite a clear dependence on semantic information. At the semantic level, using English lexical stimuli in BERT \cite{devlin-etal-2019-bert}, \citet{misra-etal-2020-exploring} found that BERT does display evidence of sensitivity to semantic priming, though this is localized to more unconstrained contexts and only certain semantic relations. 

In analyses of LLMs' linguistic understanding modeled on psycholinguistic experiments, however, the question of discourse knowledge remains relatively under-explored. Recent existing work has presented partially conflicting accounts in this area, in particular with regard to how LLMs may or may not exhibit human-like biases in pronoun interpretation. For example, \citet{davis2020discourse} compared LSTM LMs and Transformer LMs behaviors and internal representations in dealing with implicit causality verbs, finding that surprisingly (contrary to humans) implicit causality only influences Transformer LMs' behavior for reference, but influences neither model for syntactic attachment. \citet{sorodoc-etal-2020-probing} also compared LSTM LMs and Transformer LMs in coreference resolution corpora, finding that although LMs are much better at grammar, they also captured referential aspects to some extent. 

\begin{figure*}[t]
  \centering
  \includegraphics[width=\linewidth]{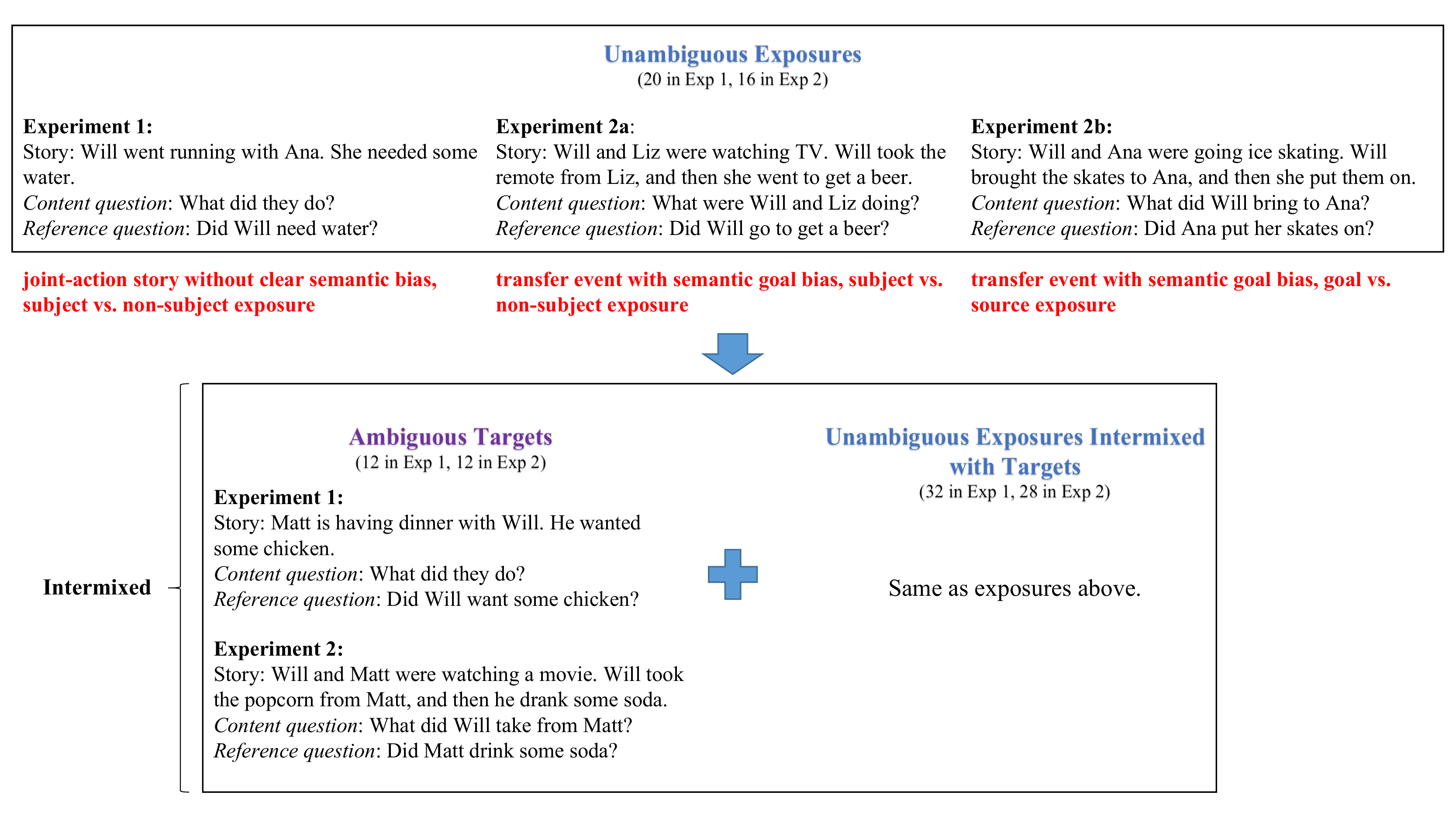}
  \caption{Illustrated Experimental Procedure. Closely following the setup with human participants in \citet{johnson2022frequency}, LLMs were primed via exposure to story/question pairs with unambiguous referents, and tested for their responses on ambiguous target pairs.}
  \label{fig:1}
\end{figure*}

This existing work replicates sets of individual stimuli from psycholinguistic experiments in isolation; by contrast, our work takes a more behaviorally-oriented approach by replicating the stimuli, procedure, and even experimental design of a full set of human psycholinguistic studies.
We do this to ask a further question: beyond exhibiting baseline human-like referential biases in pronoun interpretation, do LLMs display adaptation to the frequency of referential patterns in the local context like humans? This question is important because adapting to a referential bias over the course of an experiment requires a sustained representation of the frequency of the pattern, which involves a higher-level understanding of the discourse structure. Humans have shown abilities of this kind of adaptation at multiple linguistic levels. For instance, exposure to syntactic structures consecutively affects humans' choice of structures in both the short term and long term via priming effects (e.g., \citealt{branigan2005priming, chang2000structural}). Exposure to phonological patterns can also guide humans' segmentation patterns (e.g., \citealt{saffran1996statistical,saffran1996word}). At the semantic and pragmatic levels, humans can also adapt as listeners to speakers' variable choices of uncertainty expressions \cite{schuster2020know}.

\section{Methods}

In this work, we aim to replicate three experiments from \citet{johnson2022frequency}, transferring their designs, procedures, and stimuli as faithfully as possible to the LLM context using in-context learning.

\subsection{Source Experiments}

We first briefly summarize the experimental setup employed by \citet{johnson2022frequency}. In each experimental setting, participants heard a series of two-sentence stories in which the first sentence contained two characters with gender-marked first names (Matt or Will for men, Liz or Ana for women). For each story, participants answered a content question to check comprehension, and then a reference question to check pronominal interpretation. In order to lower the ceiling and keep participants from falling into a pattern of simply answering "yes," reference questions were equally split between default and non-default phrasings (e.g. between the subject/non-subject interpretations in Experiment 1a and 2a and between the source/goal interpretations in Experiment 2b). Figure \ref{fig:1} illustrates sample stimuli and the procedure of the three experiments conducted. 

In each experiment participants were first shown a series of stories with \textit{exposure} reference questions to establish a referential pattern; in these, the characters had different genders so pronoun interpretation was unambiguous. After the exposure phase (20 stories in Experiment 1, 12 stories in Experiment 2), further stories with unambiguous exposure questions were intermixed with 12 stories accompanied by \textit{critical} reference questions; in these stories, the characters had the same gender, so pronoun interpretation was ambiguous and required reliance on discourse cues.

\begin{figure*}[h] 
  \centering
  \includegraphics[width=\linewidth]{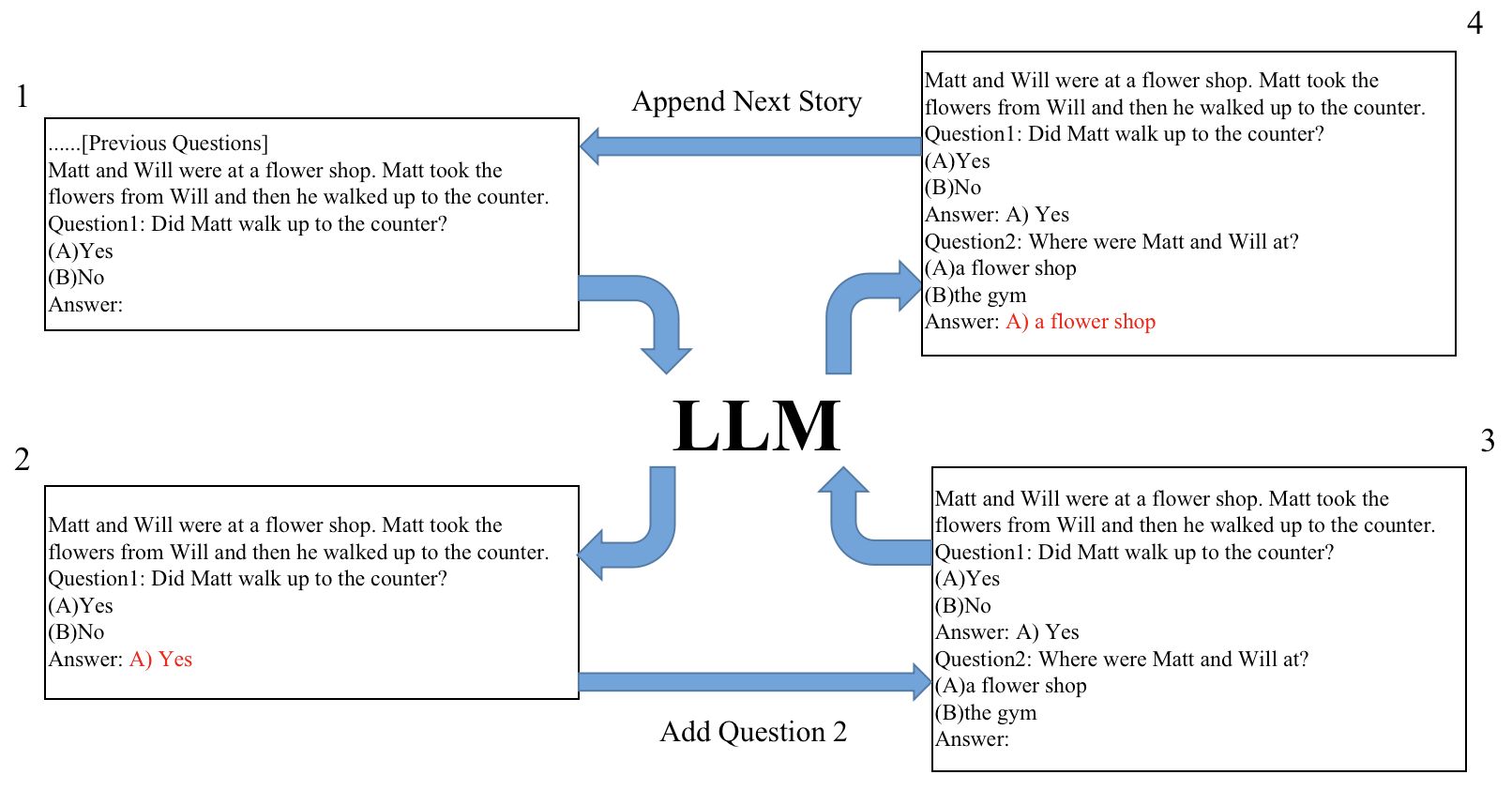}
  \caption{The ICL Simulation Framework. Black text is provided to the models as prompts, {\textcolor{red}{red text}} is generated by the model. With the last step's output appended to previous prompts, we ask LLMs the next question.}
  \label{fig:2}
\end{figure*}

Across experiments, there were two key conditions that were manipulated for exposures. Under a \textit{subject exposure} condition, the unambiguous intended referents of all exposure questions are subjects of the preceding clauses; in the corresponding \textit{non-subject exposure} condition, they are the objects of the preceding clauses. Similarly, under a \textit{goal exposure} condition the unambiguous intended referents of all exposure questions are goal referents while in a \textit{source exposure} condition they are source referents. 

Aiming to transfer these experiments as faithfully as possible to the LLM context, we used this experimental paradigm to evaluate the LLM by providing the model with the full text of each story prompt and content/reference question, then generated tokens in response which we interpreted as answers.

For clarity, we use the same experiment numbers and identifiers as in \citet{johnson2022frequency}. Note that we did not replicate experiments 1b and 1c, which investigated whether people are still sensitive to referential patterns using different types of referential expressions (e.g., third-person names and first- and second-person pronouns). While these experiments provided insights into the linguistic structure at which people generalize referential biases, they were less relevant to the objective of this study and thus were not included. 


\paragraph{Experiment 1a} In this experiment, all story prompts contain "joint-action" verbs using "with" in the form "X did something with Y." Since these verbs lack a clear semantic bias (e.g., \citealt{arnold2018linguistic}), this context allows us to evaluate LLMs' sensitivity to syntax-based biases in discourse by asking whether exposure to only subject-bias or object-bias examples will influence following answers on the ambiguous critical items. If LLMs are sensitive to syntax-based referential patterns, we expect more subject responses under the subject exposure condition and more non-subject responses under the non-subject exposure condition.

\paragraph{Experiment 2a} Experiment 2a forms a bridge between adaptation to syntactic and semantic constraints. Are LLMs able to track patterns in both categories, for instance learning an exposure bias in one category while ignoring variation in the other? In this experiment, all story prompts contain source/goal verbs like "give" and "receive," but these are distributed equally throughout exposures. The manipulation remains the same as Experiment 1a, in which LLMs are exposed to consistent and unambiguous subject interpretations in the subject exposure condition and non-subject interpretations in the non-subject exposure condition.

\paragraph{Experiment 2b} This experiment focuses solely on source/goal biases, in which all story prompts contain source/goal words, but the unambiguous exposure items are manipulated to contain only source references in the source exposure condition and only goal references in the goal exposure condition.

\subsection{In-context Learning}\label{section:icl}

We propose that since these experiments rely on short-term learning effects of exposure in an experimental context, they can only be effectively simulated with LLMs by using in-context learning recursively. Specifically, for each question, the model is provided access to all previous items in answering a new question, including its own previous responses. This is intended to mirror the process of human experimental participants making judgments in the light of recent exposure to input and their own past responses.

We manually checked the correctness of LLM responses to content questions intended to check comprehension, and as in the human experimentation context removed answers for which the LLMs provided incorrect answers. The answers to critical target questions were recorded for further statistical analysis. Our ICL procedure is shown in Figure \ref{fig:2}.

\subsection{Models and Experimental Settings}
We used \emph{text-davinci-003} from the OpenAI API and open-sourced FLAN-UL2 as the LLMs of interest. Though these models are of course not exhaustive of the current landscape of LLMs, they provide some diversity since they differ in both structure (decoder-only vs. encoder-decoder) and parameter count (175B vs. 20B). To introduce more randomness into the experiment and allow enough sample sizes for statistical comparisons across conditions, we made slight modifications to the \emph{temperature} hyperparameter on each run. Specifically, we assigned each run a random and unique temperature value between 0.2 to 1.0.

We also attempted to simulate `participants' using a natural language prompt to approach different speaker identities following a similar methodology to \citet{https://doi.org/10.48550/arxiv.2208.10264}. We developed prompts with slots for titles, names, and country of origin to establish different character backgrounds simulating native English speakers from the United States, Britain, and Australia, following the participants' demographics in \citet{johnson2022frequency}. However, these prompts did not induce greater diversity in responses than temperature modification, so only results using temperature modification are presented below. We present a further analysis for both methodologies in Appendix~\ref{sec:appendix2}.

In the end, we simulated 24 `participants' each in Experiment 1a and Experiment 2a, and 60 `participants' in Experiment 2b. We included more in Experiment 2b because this experiment has a lower response variability and thus needs more data points for the statistical analysis.


\subsection{Measures}
\begin{figure*}[h] 
  \centering
  \includegraphics[width=\linewidth]{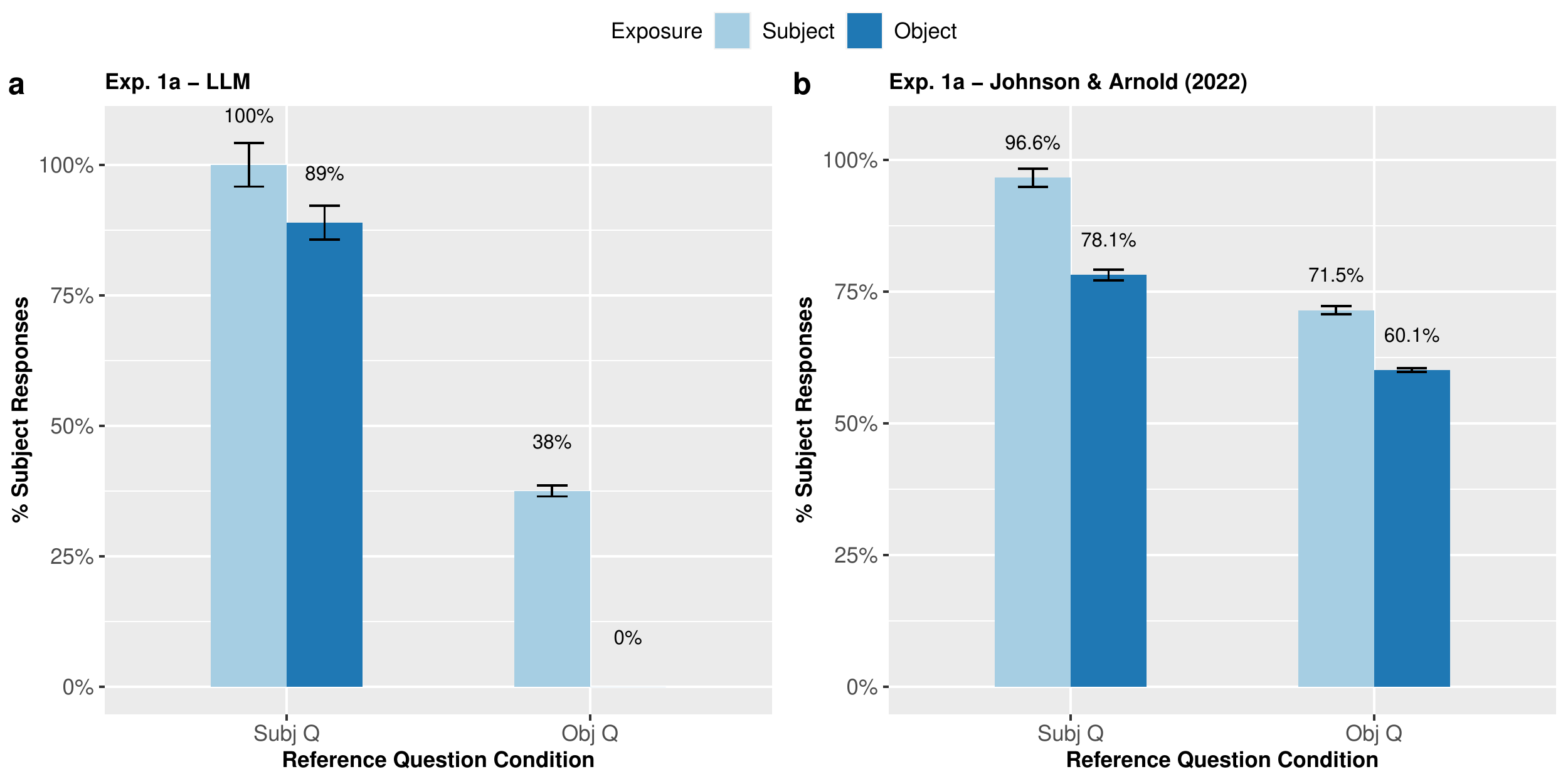}
  \caption{Subject responses of \textbf{LLM (left)} and \textbf{human participants (right)} in Experiment 1a, showing the percentage of subject responses for each type of reference question (Subj Q: subject reference question; Obj Q: object reference question), grouped by exposure type (subject exposure: light blue; object exposure: dark blue).}
  \label{fig:3}
\end{figure*}

Following the analytic approach of \citet{johnson2022frequency}, we used a regression-based approach to analyze whether the responses of LM are consistent with the subject or gogoalal bias of the context for each experiment. We can then compare our findings with theirs by asking whether the effect of the main predictors is the same. In all experiments, predictors included exposure type (unambiguous exposures to subject/non-subject or goal/source), reference question type (whether the reference question is asked about the subject/non-subject or goal/source), and the interaction effect between them. Experiment 2 included verb type, as well as its two-way and three-way interaction effects with the other two predictors as additional predictors. Given our small dataset, the results were analyzed using Bayesian mixed-effects Bernoulli logistic regression models in the R package \emph{brms} \cite{burkner2017brms} instead of a frequentist model. We report a Bayesian equivalent p-value (p\_MAP) computed with the R package \emph{bayestestR} \cite{makowski2019bayestestr} to offer a straightforward interpretation of the results. Details of models are provided in Appendix 1.

\section{Results}
\subsection{FLAN-UL2}
We found that FLAN-UL2 was not capable of generating meaningful output for analysis under this design. First, FLAN-UL2 showed a much higher false rate in answering content questions versus InstructGPT: while InstructGPT replied 100\% correctly to these questions, FLAN-UL2 replied to only 57\% correctly. Second, for the ambiguous target items, FLAN-UL2 answered `yes' 100\% of the time, indicating an extremely strong bias towards simply answering `yes' and producing no meaningful variation. By contrast, InstructGPT answered `yes' to target questions 68\% of the time, suggesting some amount of `yes' bias but to a much weaker degree. From these findings we concluded that FLAN-UL2 did not produce sufficiently clean outputs for analysis; therefore, in the following sections, we will focus on results from InstructGPT.
\subsection{Experiment 1a}
\begin{figure*}[t] 
  \centering
  \includegraphics[width=\linewidth]{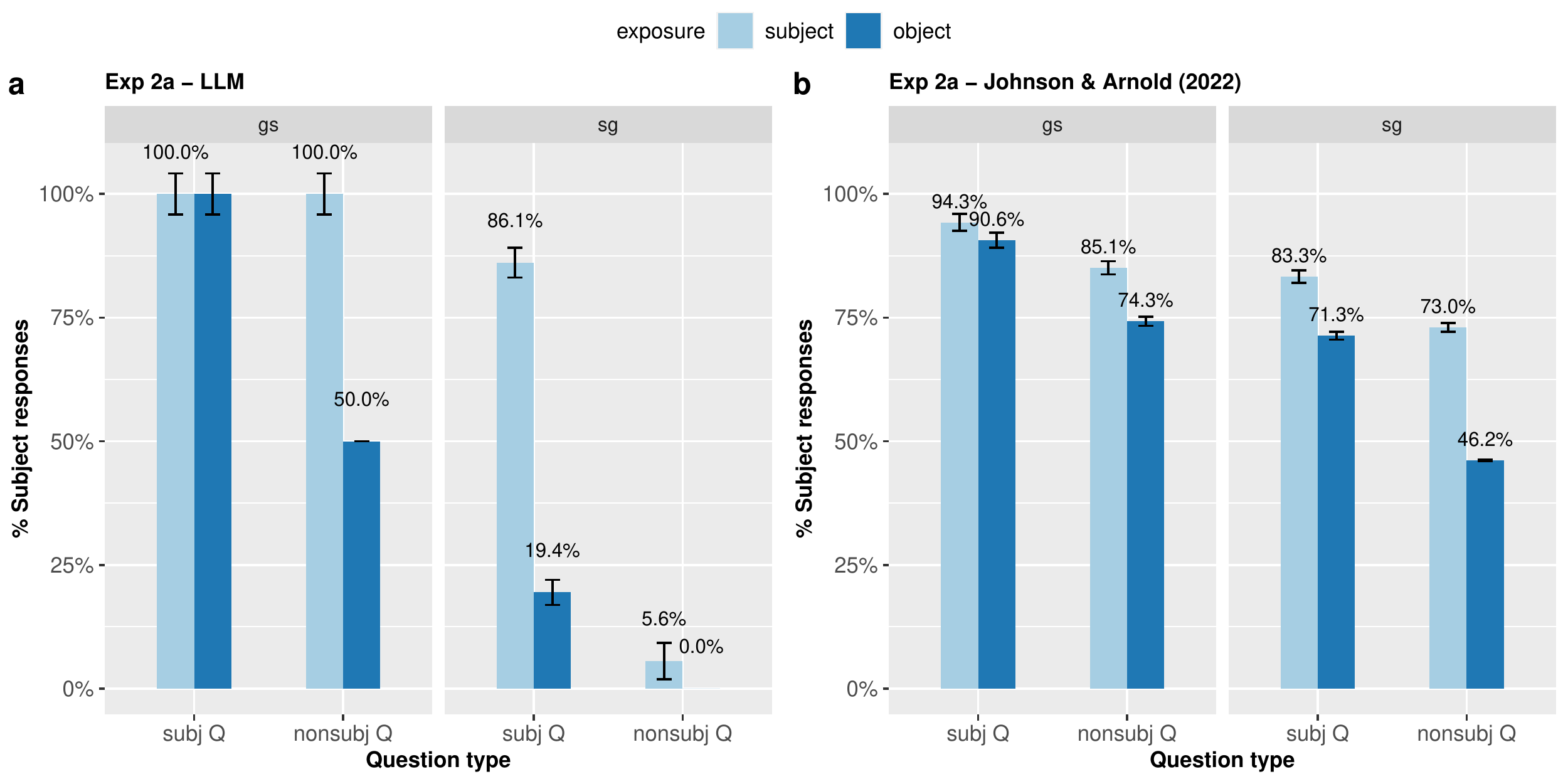}
  \caption{Subject responses of \textbf{LLM (left)} and \textbf{human participants (right)} in Experiment 2a. The proportion of subject responses is plotted against question type faceting by verb type, comparing goal-subject verbs (gs) like "receive," where the subject is the goal referent) to subject-goal verbs (sg) like "send," where the subject is the source referent.}
  \label{fig:4}
\end{figure*} 
\begin{figure*}[t] 
  \centering
  \includegraphics[width=\linewidth]{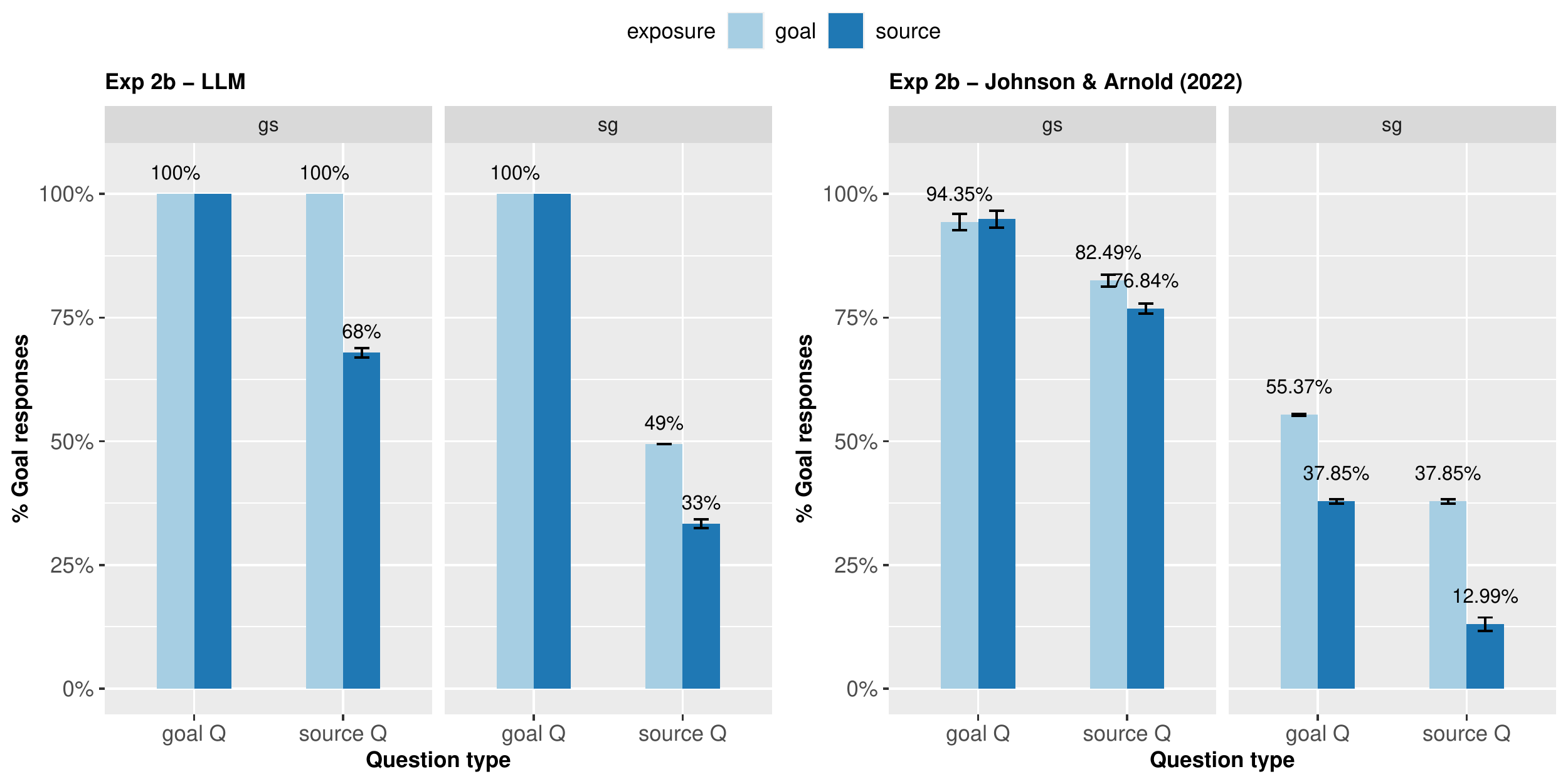}
  \caption{Goal responses of \textbf{LLM (left)} and \textbf{human participants (right)} in Experiment 2b.}
  \label{fig:5}
\end{figure*}

Experiment 1a asks whether LLMs are sensitive to the frequency of referential patterns when subject referents are preferred. Figure \ref{fig:3} compares the subject responses of our results with \citet{johnson2022frequency}'s results. In both LLM and human data, we saw fewer subject responses in object exposure than in subject exposure. This is confirmed by the main effect of exposure type (\textit{p\_MAP} < .001) revealed in the statistical analyses. There was also a main effect of response question type (\textit{p\_MAP} = .009). As seen from the figure, there were more subject responses in the subject than object reference questions in both human and LLM data. This is because InstructGPT did answer `yes' more frequently in general, which led to more subject interpretations when the question asked about the subject (where answering `yes' indicates a subject interpretation), and fewer subject interpretations when the question was asked about the non-subject (where answering `yes' indicates a non-subject interpretation).

However, we did not find any interaction effect between exposure type and reference question type. While \citet{johnson2022frequency} found that exposure type has a significant effect for subject-referent questions but only a marginal effect for the nonsubject-referent questions, the exposure effect was significant for both question types in our results. InstructGPT is sensitive to exposure type even for nonsubject-referent questions, as reflected in Figure \ref{fig:3}: there were no subject responses when non-subject reference questions were asked under the non-subject exposure condition. This suggests that LLM may even be more sensitive to the exposure effect than humans. Overall, Experiment 1 shows that LLM can indeed learn from recent exposure to syntactically-oriented referential patterns, 
in a relatively more non-subject-biased way.

\subsection{Experiment 2a}


We 
ask in Experiment 2a if the sensitivity to subject reference patterns observed in Experiment 1a persists independent of semantic variability in the local context. Figure \ref{fig:4} illustrates the responses of LLM and human participants by exposure, reference question, and verb type. The behavior of InstructGPT is in line with what \citet{johnson2022frequency} found in human participants for Experiment 2a: as in Experiment 1, statistical analyses revealed a significant main effect of referent question type (\textit{p\_MAP} = .034), despite a marginally significant effect of exposure type (\textit{p\_MAP} = .085). InstructGPT did understand the pronoun as subject referents more after subject exposure and said `yes' more often such that there were more subject interpretations when the critical question was asked about the subject. InstructGPT also showed a goal bias: there was a main effect of verb type (\textit{p\_MAP} = .001), such that it referred to the subject more with gs verbs (where the subject is the goal referent) than sg verbs (where the subject is the source referent). We observed no interaction effect. 

In spite of these similarities, we still observe differences between LLM and human participants. Notice in Figure \ref{fig:4} that there is a larger difference between gs and sg verbs in InstructGPT than in human participants while keeping other conditions constant, suggesting that LLMs may have a larger goal bias than humans.

\subsection{Experiment 2b}
Experiment 2b examines whether InstructGPT is still sensitive to exposure to referential patterns that exhibit consistent preferences for a source or goal referent. Figure \ref{fig:5} compares the results from InstructGPT and human participants. Whereas \citet{johnson2022frequency} reported significant main effects from exposure type, reference question type, and verb type, as well as a marginal interaction effect between verb type and exposure, we did not find any effect from these predictors, but only an interaction effect between verb type and question type (\textit{p\_MAP} < .001). We further examined the effect of question type and exposure type for each verb type, but no significant effect was found for either predictor. These results suggest that exposure did not necessarily change InstructGPT's behaviors: for goal questions, both the goal and source exposure conditions led to 100\% goal responses. Reference question type also did not affect InstructGPT's responses, because it almost always interprets the pronoun as the goal referent under subject exposure. The lack of any significant effects, and indeed our observed universal goal interpretations to goal-focused reference questions, suggest that InstructGPT displays an extreme goal bias in pronoun interpretation.

\section{Discussion}
The present study examined whether and to what extent LLMs display adaptation for pronominal interpretation after exposure to referential patterns by replicating three psycholinguistic experiments. Exposed to the same stimuli and study design, and analyzed with the same statistical procedures as the source experiments, we asked whether LLMs show human-like behaviors and compared the performance of LLMs-simulated participants with human participants. 

We firstly found a difference between the capacity of contemporary models to replicate human psycholinguistic experimental designs in an ICL framework. While InstructGPT was able to correctly answer all the presented comprehension-check content questions, FLAN-UL2 was not. Both models displayed at least some bias towards answering `yes' in ambiguous cases, but in the case of FLAN-UL2 this bias resulted in 100\% `yes' answers, rendering meaningful variation impossible to ascertain. This could be a result of structural differences between the models (decoder-only in the case of InstructGPT, encoder-decoder in the case of FLAN-UL2), or perhaps more likely a question of simple model size (175B for InstructGPT vs. 20B for FLAN-UL2).

Experiments 1a and 2a examined whether LLMs adapt their syntactic bias from recent exposure to referential patterns like humans, without and with the presence of possibly confounding semantic goal bias. We found that LLM's referential biases are indeed sensitive to such exposure in both experiments. In addition, LLM did exhibit a goal bias in Experiment 2a, in line with previous studies which argue for LLM's ability to exhibit human-like semantic bias in pronoun interpretation (e.g., \citealt{davis2020discourse}).

Experiment 2b examined whether LLM can adapt and vary their semantic bias from exposures. In this context, in contrast with the previous two experiments, exposure type did not affect LLM behavior at all. This raises the question of why LLM would be sensitive to exposure to only referential patterns that exhibit consistent syntactic bias but not semantic bias. An immediate and intuitive answer would be that LLM is unable to fully represent semantic knowledge in referential processing. However, given that LLM did display a human-like goal bias in pronoun interpretation independent of exposure, this explanation seems unlikely. 

We suggest that our work provides evidence that LLM only partially represents the semantic knowledge involved in referential processing for two reasons. 

First, adapting referential biases on the basis of exposure in the local context may require representations of higher-level knowledge than merely exhibiting a bias towards certain referents. While the latter may only require knowledge of which features associated with a referent are more frequent or likely in general, adaption requires a sustained awareness of referential pattern frequency as it changes in the local discourse context. Though representing knowledge about semantic relations was observed as early as the analogical reasoning task in Word2Vec \cite{NIPS2013_9aa42b31}, human-like extraction of abstract information like a persistent discourse state is more challenging. The model may only be able to identify thematic roles (source or goal) of a referent and associate them with pronoun interpretation, but not to identify a consistent pattern of thematic roles across a discourse. If so, this would explain the strong goal bias we observed in Experiment 2. 

Second, it is possible that LLM has an extremely strong goal bias that masks the influence of exposures. If so, this suggests that LLM over-represents the semantic knowledge encoded in pronoun interpretation. 

In either case, our results suggest that LLM do not encode semantic knowledge in a fully human-like way, even though they do demonstrate some human-like capacities for semantic understanding. 
Although we believe this gap can be mitigated via instruction fine-tuning or chain-of-thought prompting \cite{https://doi.org/10.48550/arxiv.2201.11903}, these results still suggest we should consider incorporating semantically-informed objectives into self-supervised pre-training to a greater extent.

\section{Conclusions and Future Work}
By replicating a series of psycholinguistic experiments as closely as possible using in-context learning, this paper pioneered whether LLMs would adapt pronominal interpretation behaviors in a human-like way given exposure to referential patterns in the local discourse context. Our work suggests paths forward for replicating psycholinguistic experiments in a more faithful way that allows for comparisons between human and LLMs' behaviors. 


\section*{Limitations}
Several main limitations exist in our study in its current form. First, our reported results only simulated experimental participants by manipulating the \emph{temperature} hyperparameter. We compared this approach with natural language prompting for Experiment 1, but that prompting did not increase "participant" diversity, so it was abandoned. Moreover, approaches for simulating psycholinguistic experimental "participants" could go far beyond what was tried here; our prompting method was relatively limited, and more detailed prompting could be included in future experimental simulations. Second, making a direct comparison with actual psycholinguistic experiments might not be the only method to investigate LLMs' discourse capacity. A comprehensive list of discourse probing tasks might play a similar role despite a different way \cite{koto-etal-2021-discourse}. Third, this study is strictly behavioral: limited by both computational resources and obscure mechanisms of in-context learning, we do not dive into models' internal representations in our analyses. 

\section*{Acknowledgement}
We are very grateful to Jennifer Arnold for sharing the stimuli and design of \citet{johnson2022frequency} with us.

\bibliography{anthology,custom}
\bibliographystyle{acl_natbib}

\appendix

\section{Statistical Information}
\label{sec:appendix1}
For each experiment, we analyzed referent choice (Subject = 1, Non-subject = 0) using a mixed-effects Bernoulli regression model from the R package \emph{brms} \cite{burkner2017brms}, with the maximal random structure justified by design \cite{barr2013random}. Predictors are coded in the same way as in \citet{johnson2022frequency}. All models were specified with a weakly informative prior using the Cauchy distribution with center 0 and scale 2.5. Models were fitted using six chains, each with 4,000 iterations of which the first 1,000 are warmup to calibrate the sampler, resulting in 18,000 posterior examples.  

The model for Experiment 1a included question type (QtypeC, sum-coded: Subject = 0.5, Non-subject = -0.5) and exposure type (PC, effects-coded: Subject-biased = 0.51, Object-biased = -0.49) as fixed predictors, random intercepts for participants and items, random slopes of question type, exposure type and their interaction for items, and a random slope of question type for the participant. Exposure type was not included as a random slope for participants because the condition does not vary within participants.
\begin{verbatim}
brm (Rc ~ QtypeC*PC +
        (1+QtypeC*PC|ID)+(1+QtypeC|Subject), 
           data=filter(e1a,Exposure!="None"),
        family="bernoulli", chains=6, 
            iter=4000, warmup=1000,
            control = list(adapt_delta = 0.95),
            prior=
            c(set_prior ("cauchy(0,2.5)")))
\end{verbatim}

The model for Experiment 2a included question type (QtypeC, sum-coded: Subject = 0.5, Non-subject = -0.5), exposure type (PC, sum-coded: Subject-biased = 0.5, Object-biased = -0.5), and verb type (Vc, sum-coded: gs-verb = 0.5, sg-verb = -0.5) as fixed predictors, random intercepts for participants and items, random slopes of question type, exposure type and their interaction for items, and a random slope of question type for the participant. As in Experiment 1a, exposure type was not included as a random slope for participants because the condition does not vary within participants. Similarly, verb bias was not included as a random slope for items here because it does not vary within items.
\begin{verbatim}
    brm (Rc ~ QtypeC*PC*Vc +
        (1+PC*Qtypec|Item)+
        (1+QtypeC*Vc|Subject), 
        data=e2a,
        family="bernoulli", chains=6, 
        iter=4000, warmup=1000,
        control = list(adapt_delta = 0.98),
        cores = 6,
        prior=
        c(set_prior ("cauchy(0,2.5)")))
\end{verbatim}

The model for Experiment 2b included question type (QtypeC, sum-coded: Goal = 0.5, Source = -0.5), exposure type (PC, sum-coded: Goal-biased = 0.5, Source-biased = -0.5), and verb type (Vc, sum-coded: gs-verb = 0.5, sg-verb = -0.5) as fixed predictors. The random effect structure was the same as that of Experiment 2a.

\begin{verbatim}
     brm (Rc ~ PC*Vc*QtypeC+(1+PC*QtypeC|Item)
        +(1+Vc*QtypeC|Subject), 
        data=e2b,
        family="bernoulli", chains=6, 
        iter=4000, warmup=1000,
        control = list(adapt_delta = 0.999),
        cores = 6,
        prior=
        c(set_prior ("cauchy(0,2.5)")))
\end{verbatim}


\section{Temperature vs. Prompt}
\label{sec:appendix2}
Our prompt-based simulation of multiple participants embedded names, countries, prefixes, and genders into a carrier sentence: \emph{\{Prefix + Name\} is a native English speaker living in \{Country\}. \{Gender\} is asked in a psycholinguistic experiment to answer the following questions.} For example, \emph{Mr. Smith is a native English speaker living in England. He is asked in a psycholinguistic experiment to answer the following questions.}

Specifically, we calculated the variance and ran the Levene's test for any significant difference between humans and InstructGPT in each experiment. In experiment 1, the human responses’ variance is 0.055, the temperature-based responses’ variance is 0.024, and the prompt-based responses’ variance is 0.017. Human responses were significantly higher than both (Temperature-based: \textit{p = .049}; Prompt-based: \textit{p = .007}). Yet, temperature-based were not significantly higher than prompt-based. Due to the limitation of API pricing, we only ran temperature-based in the following experiments. In experiment 2a, the human responses’ variance is 0.034 and the temperature-based responses’ variance is 0.043. Yet, Levene's test did not reveal any significant difference. In Experiment 2b, the human responses’ variance is 0.020 and the temperature-based responses’ variance is 0.008 (\textit{p < .001}).

We used different techniques to introduce randomness and include more experimental data in our experiments. We realized these were not well-designed prompts to elicit different linguistic backgrounds. Given the lack of investigation on simulating multiple participants in psycholinguistics studies, we recognize this as a future direction of possible work.

\end{document}